%% file: wacv_title3-copy.tex
\documentclass[10pt,twocolumn,letterpaper]{article}

\usepackage{wacv}
\usepackage{times}
\usepackage{epsfig}
\usepackage{graphicx}
\usepackage{amsmath}
\usepackage{amssymb}

\graphicspath{{figs/pdf/}}
\usepackage{cleveref}
\crefname{subfigure}{fig.}{fig.}

\def\eg{\textit{e.g.}}
\def\ie{\textit{i.e.}}

\def\etal{\textit{et~al.}}

\def\feat#1{\mathcal{F}_{\text{#1}}}

\usepackage[pagebackref=true,breaklinks=true,letterpaper=true,colorlinks,bookmarks=false]{hyperref}

\wacvfinalcopy 

\def\wacvPaperID{96} 

\ifwacvfinal\fi
\begin{document}

\title{Finding Temporally Consistent Occlusion Boundaries \\ in Videos using Geometric Context}



\author{S. Hussain Raza$^{1,3}$ \hspace{.25cm} Ahmad Humayun$^{2}$ \hspace{.25cm} Matthias Grundmann$^{4}$ \hspace{.25cm} David Anderson$^{1}$ \hspace{.25cm} Irfan Essa$^{1,2}$\\[.1in]
$^{1}$School of Electrical and Computer Engineering, Georgia Institute of Technology, Atlanta, GA, USA\\
$^{2}$School of Interactive Computing, Georgia Institute of Technology, Atlanta, GA, USA\\
$^{3}$Nvidia Corporation, Santa Clara, CA, USA\\ 
$^{4}$Google Research, Mountain View, CA, USA\\
{\tt\small \href{http://www.cc.gatech.edu/cpl/projects/temporaloccl}{http://www.cc.gatech.edu/cpl/projects/temporaloccl}}
}

\maketitle
\ifwacvfinal\thispagestyle{empty}\fi

\input{abstract}
\input{intro}

\input{related}

\input{dataset}

\input{method}

\input{mrf_occl}


\input{results}

\input{conclusion}

{\small
\bibliographystyle{ieee}
\bibliography{egbib}
}

\end{document}

%% file: abstract.tex
\begin{abstract}
We present an algorithm for finding temporally consistent occlusion boundaries in videos to support segmentation of dynamic scenes. We learn occlusion boundaries in a pairwise Markov random field (MRF) framework. We first estimate the probability of an spatio-temporal edge being an occlusion boundary by using appearance, flow, and geometric features. Next, we enforce occlusion boundary continuity in a MRF model by learning pairwise occlusion probabilities using a random forest. Then, we temporally smooth boundaries to remove temporal inconsistencies in occlusion boundary estimation. Our proposed framework provides an efficient approach for finding temporally consistent occlusion boundaries in video by utilizing causality, redundancy in videos, and semantic layout of the scene. We have developed a dataset with fully annotated ground-truth occlusion boundaries of over $30$ videos ($\sim$5000 frames). This dataset is used to evaluate temporal occlusion boundaries and provides a much needed baseline for future studies. We perform experiments to demonstrate the role of scene layout, and temporal information for occlusion reasoning in dynamic scenes.\end{abstract}

%% file: intro.tex
\section{Introduction}
\label{sec:intro}

\noindent Objects in a scene exhibit occlusion due to their depth ordering with respect to the camera. In video, occlusion relationships can change over time due to ego-motion or movement of the objects themselves. In both cases, edges of the objects give occlusion boundaries. These occlusion boundaries are a strong indicator of object segmentations. Hoiem~\etal\cite{hoiem2007occl} showed that by reasoning about occlusions, object segmentation, recognition, and scene description in images can be improved. To locate these edges, some initial estimates of motion and segmentations are required, but typical algorithms tend to fail close to these boundaries due to depth inconsistency. In this paper, we estimate these occlusion boundaries by combining low level appearance and flow cues with higher level information like geometric scene labels. These estimates of boundaries provide significant improvements to spatio-temporal video segmentation. 

\begin{figure}
\centering
\includegraphics[width=\columnwidth]{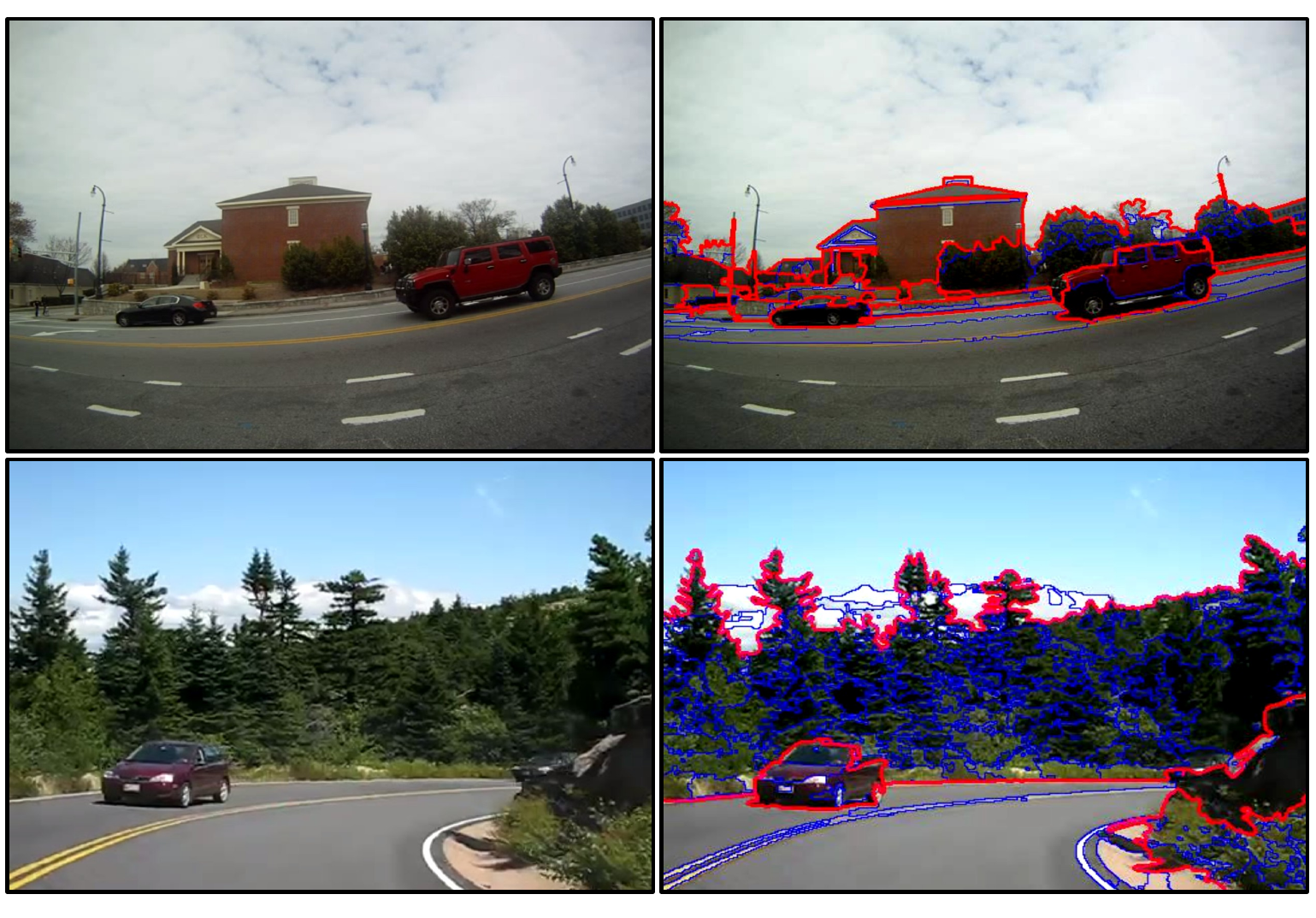}
\caption{{Video frames of an urban scene, occlusion and non-occlusion boundaries are labeled as red and blue, respectively. We demonstrate importance of geometric features and temporal redundancy for finding temporally consistent occlusion boundaries. }}
\label{fig:intro}
\end{figure}



Our algorithm learns temporally consistent occlusion boundaries in dynamic scenes by leveraging spatio-temporal segmentation of videos. We first segment a video into spatio-temporal super-voxels \cite{MatthiasSegmentation,XuXiCoECCV2012}. Over-segmentation provides a large number of candidate boundaries for learning occlusion/non-occlusion boundaries. We extract a broad range of features from each segment's boundary, and train unary and pairwise boundary classifier and enforce occlusion boundary continuity in MRF. MRF enables us to encode pairwise edgelet relations into our model, \ie, probability of an occlusion boundary to be connected to other occlusion and non-occlusion boundaries, reducing false positives. We also demonstrate that aggregating information about occlusion boundaries over a temporal window increases performance when compared to a frame by frame approach. For testing and evaluations, we have developed a large dataset consisting of outdoor videos, annotated with occlusion boundaries.
\vspace{-0.5cm}
\paragraph{Our primary contributions are:}  (1) a method for estimating temporally consistent occlusion boundaries by combining appearance, flow, and semantic scene information in an MRF framework; (2) a thorough evaluation of our algorithm by examining feature importance in estimating occlusion boundaries and comparison with other occlusion boundary algorithms (see \Cref{sec:results}); (3) in addition, we introduce a  novel dataset of 30 annotated videos ($\sim$5,000 frames) with temporal occlusion boundaries and semantic information, as existing datasets do not provide temporal and semantic annotations.


%% file: related.tex
\vspace{-0.2cm}
\section{Related Research}
\label{sec:related}
\vspace{-0.1cm}
Geometric layout and temporal consistency in a dynamic scene provide strong cues for scene understanding and object segmentation. Hoiem \etal\cite{hoiem2007occl} demonstrated importance of geometric features for occlusion detection for images. Saxena~\etal\cite{saxena2009make3d} proposed a planar model for estimating 3D structure of the scene from a single image. Applying image-based methods to individual video frames can provide occlusion reasoning of the dynamic scene. However, such image-based methods may not exploit the temporal information across frames, leading to temporally inconsistent scene description. 

Detecting occlusion boundaries is a well studied problem, due to its usefulness in understanding the depth, motion and context of the scene~\cite{stein2009occlusion,humayun2011learning}. Fleet~\etal~\cite{fleet2003bayesianmotionboundaries} gave a Bayesian formulation where boundaries resulted from distinguishing local image motion. Stein~\etal~\cite{stein2009occlusion} has shown that combining appearance and motion cues improves occlusion boundary detection. They further improve occlusion boundary detection by applying a global conditional random field where the potentials are learned from AdaBoost. He \etal\cite{he2010occlusion} showed that a global model may not be necessary for sequences with ego-motion and achieved comparable results by local edge and psuedo-depth maps. Recently, Sundberg~\etal~\cite{sundberg2011occlusion} improved over these boundaries by computing motion gradients across static boundaries. Since these methods rely on local features they are unable to reduce false positives where intra-object local motion or appearance variance is high. Typical examples include waves in the water or trees in the wind. In our method, semantic/geometric labels help suppress such errors.

Other methods have also been proposed to detect occlusion boundaries in a single image. Many methods inferring geometric labels initially estimate boundaries in single images~\cite{saxena2009make3d,saxena-ijcv2008,gould2009semantic}. Probabilistic boundary detectors like Pb~\cite{martin_PAMI_2004_pbedge} use local oriented energy, color, and texture gradients. Arbel\'{a}ez~\etal~\cite{arbelaez2011gpb} improve boundary detection by imposing global constraints via spectral clustering which results in closed contours. Leordeanu~\etal~\cite{leordeanu2012efficient} proposed \emph{Gb}, which reduces the time for generalized boundary detection by efficient computing boundary normals. In the last year, probabilistic boundaries have become feasible to use for realtime applications. The first method that deserves mention is Sketch Tokens~\cite{lim_CVPR_2013_sketchtokens}, which classifies edge patches using a random forest. Following this work, Doll\'{a}r and Zitnick~\cite{dollar_ICCV_2013_edge} introduced a realtime structure learning method for edge detection. From our point of view, both of these methods make many leaps forward in the single image boundary detection problem. Yet, extending these methods to videos is a non-trivial challenge. In this paper we compare to both Sketch Tokens, and the single-scale (SE-SS\_T4). and multi-scale (SE-MS\_T4) version of Structured Edges. Unlike previous methods, we use geometric or semantic labels over video segmentation for finding temporal consistent boundaries in videos.

In this paper, we leverage video segmentation to find temporally consistent boundaries in dynamic scenes. We use flow, and geometric features for estimating each edgelet's occlusion probability, and then enforce boundary continuity in a pairwise MRF framework. We demonstrate the importance of temporal smoothing and geometric features in occlusion boundary estimation. To verify our claims, we developed a comprehensive video occlusion boundary ground truth dataset with a broad set of examples.

%% file: dataset.tex
\vspace{-0.2cm}
\section{Dataset and Annotation}
\label{sec:data}
\vspace{-0.2cm}
\paragraph{Existing Datasets:}
A comprehensive dataset is necessary for evaluation of temporal occlusion boundary detection in dynamic scenes. However, existing datasets are limited to ground truth annotation for intermittent frames in a sequence~\cite{brostow2008segmentation}, and not all include semantic information. This poses a hurdle in the study of the role of the scene structure, and temporal dynamics for occlusion reasoning. Two widely used datasets for occlusion detection in videos are proposed by Stein~\etal~\cite{stein2009occlusion} and Sundberg~\etal~\cite{sundberg2011occlusion}. These datasets are limited in their scope as  (1) they provide ground-truth for only a single frame; (2) they are not suitable for the study of the role of the scene layout in occlusion reasoning and were not developed for that purpose. Butler~\etal~\cite{Butler:ECCV:2012} developed MPI-Sintel flow dataset which contains motion boundaries but does not include the occlusion boundaries in static background. The only dataset with semantic labels, and occlusion boundaries was proposed by Hoiem~\etal~\cite{hoiem2007occl}. They proposed a dataset of 50 images with ground truth annotation of outdoor scenes with occlusion boundaries, surface layout, and depth order. Since this dataset contains only a single image for a scene, it is also not ideal for our study. To overcome this limitation, we have developed a comprehensive dataset with temporal occlusion boundaries and semantic annotations.

\begin{figure}[t!]
\centering
\vspace{-0.5cm}
\includegraphics[width=0.8\columnwidth]{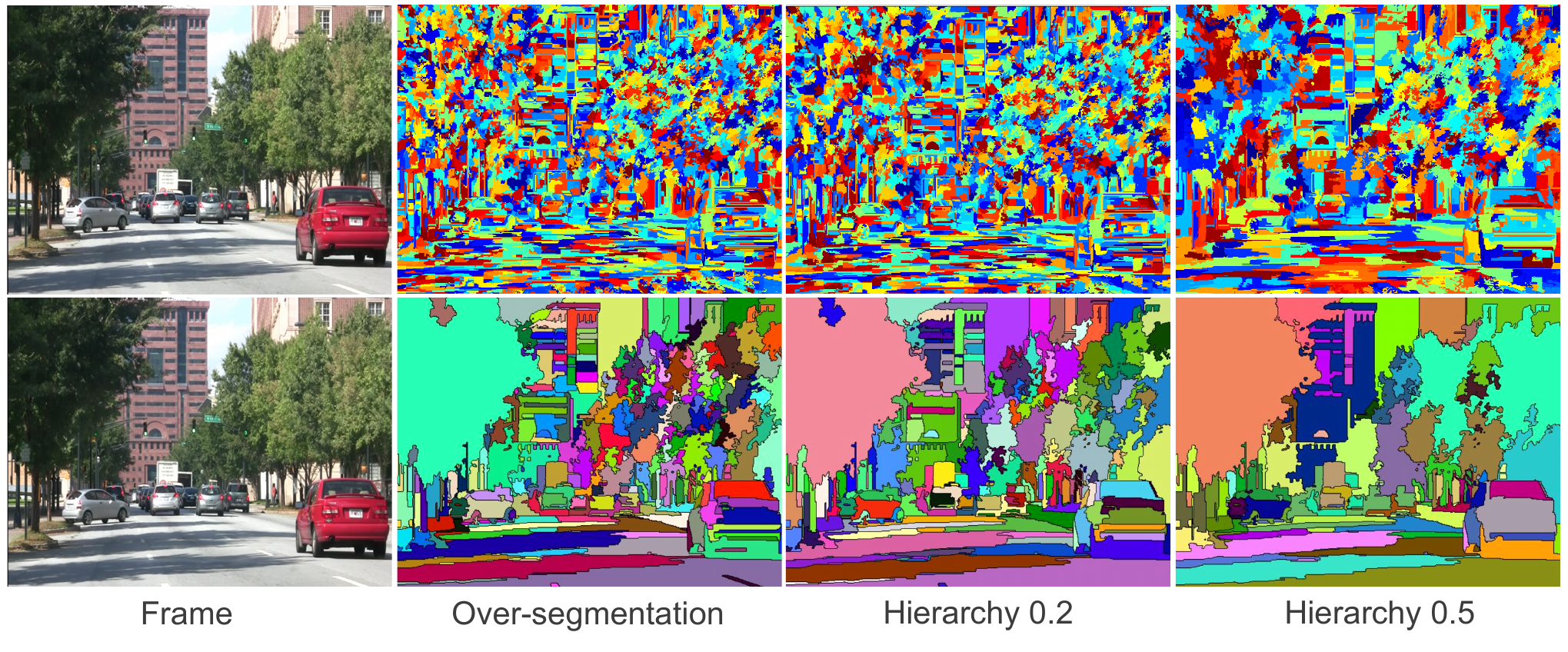}
\caption{\small{Video segmentation from Xu~\etal~\cite{XuXiCoECCV2012} at the top row and from Grundmann~\etal~\cite{MatthiasSegmentation} at the bottom row. We selected the over-segmentation (hierarchy level=0) from Grundmann~\etal~\cite{MatthiasSegmentation} because of its performance in preserving occlusion boundaries and longer temporal coherence over our challenging dataset.}}
\vspace{-0.5cm}
\label{fig:seg_compare}
\end{figure}

\vspace{-0.5cm}
\paragraph{A Video Dataset for Occlusion Reasoning:} Our dataset consists of 30 outdoor videos of urban scenes. Some videos were recorded while walking, some while driving, and others were downloaded from YouTube. We also included few videos from recently released video geometric context dataset from Raza \etal \cite{Raza2013GCFV}. The videos contain sky, ground, roads, pavements, rivers, buildings, trees, humans, and cars. Annotating temporal occlusion boundaries is a challenging task and there has been no such dataset until now. We annotate temporal occlusion boundaries in videos by using video segmentation, similar to the approach by Hoiem \etal to annotate image dataset using super-pixels \cite{hoiem2007occl}. Recently, two video segmentation algorithms have been proposed~\cite{MatthiasSegmentation,XuXiCoECCV2012}. Both these algorithms provide a hierarchy of segments from a video. We show a video and segmentation hierarchy output of these algorithms in \Cref{fig:seg_compare}. The algorithm by Xu \etal gives a  high number of super-voxels with very short temporal life, while the output from Grundmann \etal gives less super-voxels with longer temporal life as well as preserving the occlusion boundaries. We therefore selected the video segmentation algorithm proposed by Grundmann \etal \cite{MatthiasSegmentation} to annotate temporal occlusion boundaries.

\begin{figure}[t]
\centering
\vspace{-0.5cm}
\includegraphics[width=0.8\columnwidth]{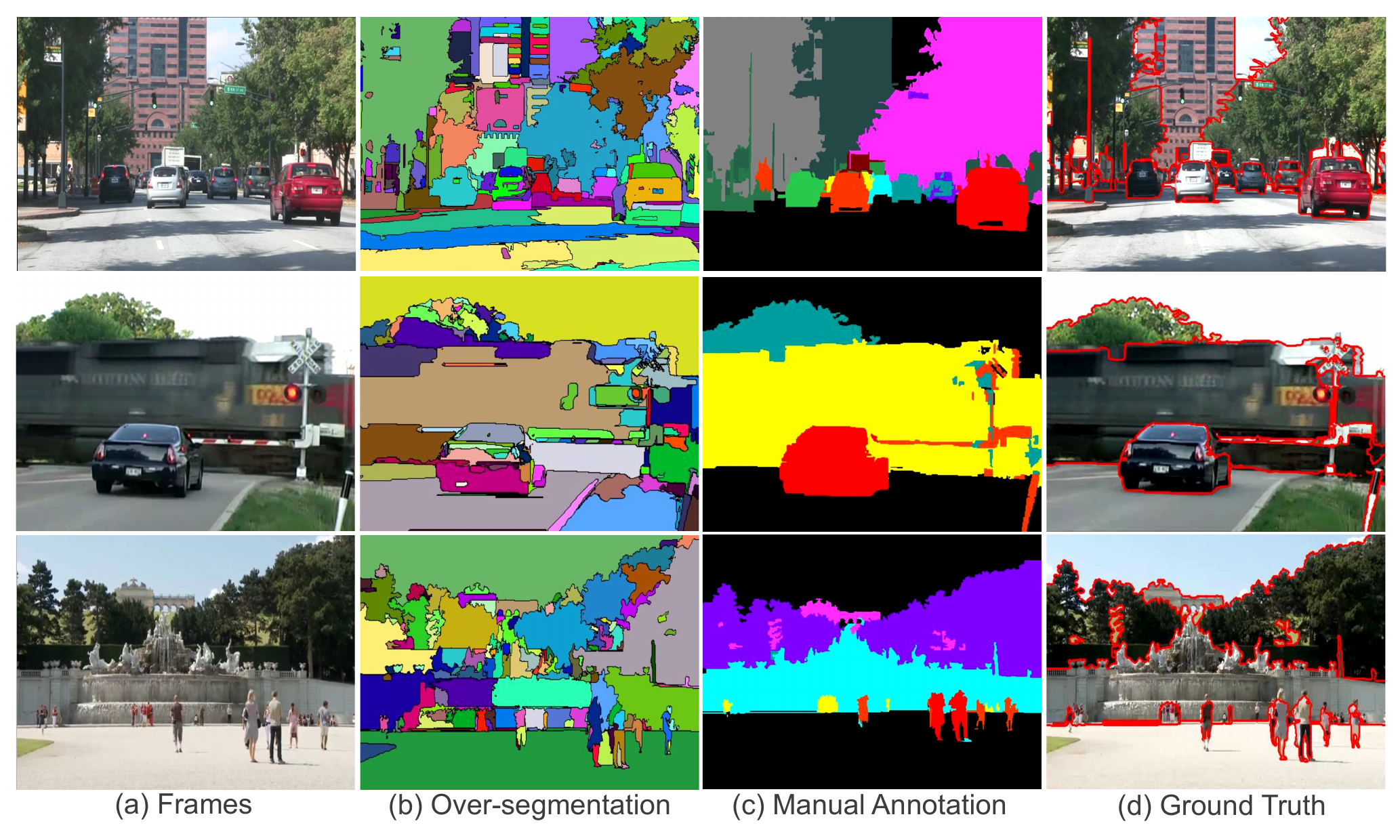}
\caption{\small{Occlusion boundary annotation: (a) an input video, (b) spatio-temporal super-voxels, (c) we cluster super-voxels into semantic classes and objects, (d) boundaries between these semantic classes and objects give occlusion boundaries.}}
\vspace{-0.5cm}
\label{fig:occl_anno}
\end{figure}

We use \cite{Raza2013GCFV,MatthiasSegmentation} to annotate the video \footnote{www.videosegmentation.com}. We group together the spatio-temporal super-voxels that belong to individual objects, as well as geometric class labels. Geometric classes annotated in our dataset are sky, ground (roads, pavements, grass, rivers),  planer surfaces (buildings and rocks), porous (trees and foliage), and movable objects (humans, cars, and trains). Boundaries between individual objects and geometric classes provide occlusion boundaries. We show the process of our manual occlusion boundary annotation in \Cref{fig:occl_anno}. In very few cases, the segmentation algorithm fails to segment a region due to similarity in color or poor-lighting condition. These boundaries are not annotated in our dataset but such cases are only a small fraction of the whole dataset. The proposed dataset contains 5,042 annotated frames across 30 videos. \Cref{Table:occl_data} provides a comparison of our dataset with existing datasets.

\begin{table}[t]
\centering
\tiny
\begin{tabular}{|l|c|c|c|}\hline
\textbf{Name} &  \textbf{Image/Video}  & \textbf{Ground-truth frames} &  \textbf{Semantic Labels}  \\\hline\hline
\emph{CMU Geometric Context}~\cite{hoiem2007occl}  & Image & 50 & Yes  \\\hline
\emph{CMU Occlusion}~\cite{stein2007combining} & Video &  30 & No\\\hline
\emph{BSDS}~\cite{sundberg2011occlusion} & Video & 60 & No \\\hline
\emph{Ours} & Video & 5042 & Yes \\\hline
\end{tabular}
\vspace{0.2cm}
\caption{Comparison with existing datasets providing ground truth for occlusion boundaries. Our dataset contains 5042 frame across 30 videos with annotations for occlusion boundaries, and geometric class labels.}
\label{Table:occl_data}
\vspace{-0.5cm}
\end{table}

%% file: method.tex
\vspace{-0.1cm}
\section{Approach}
\label{sec:method}
\begin{figure}[t]
\centering
\vspace{-0.5cm}
\includegraphics[width=\columnwidth]{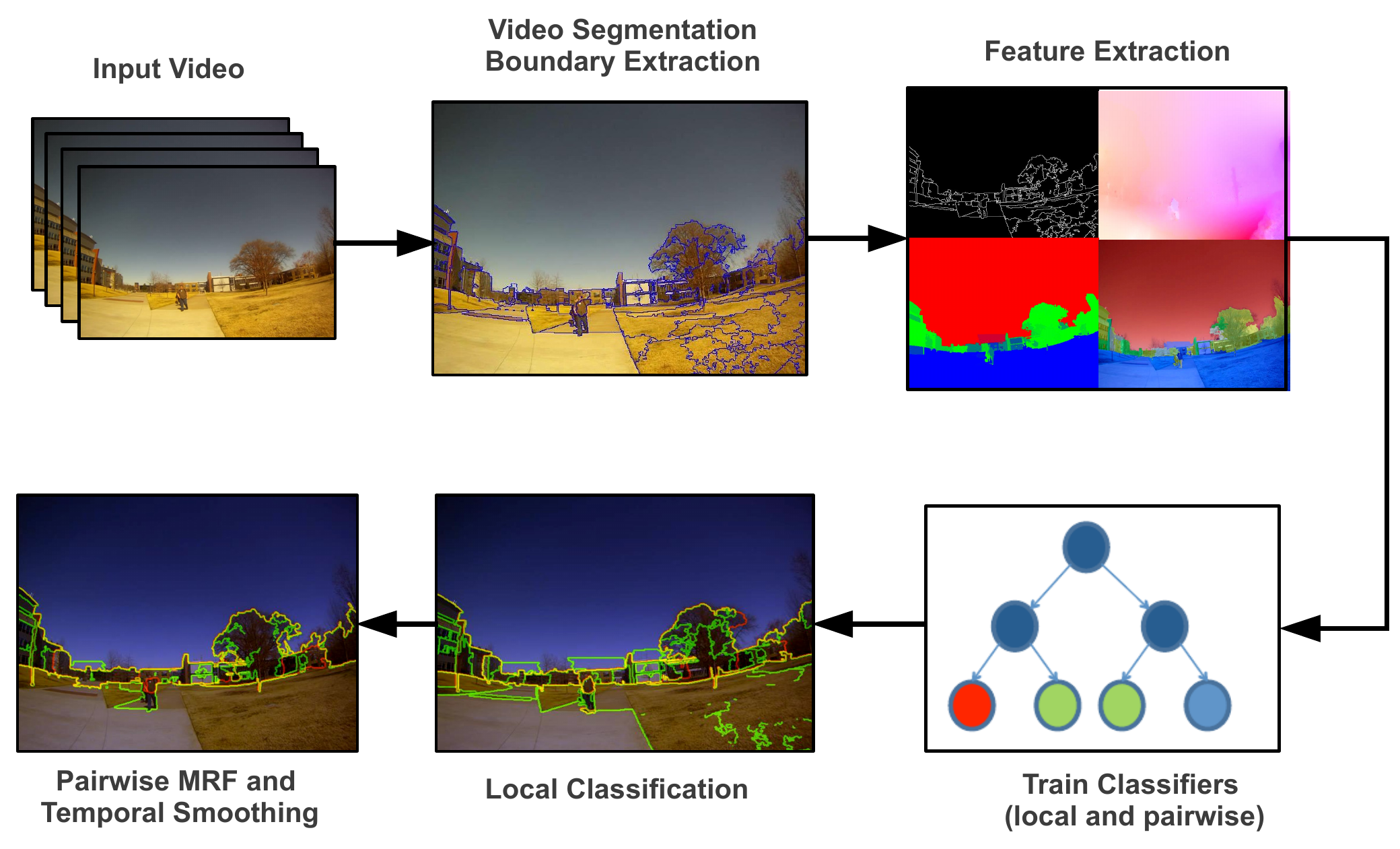}
\caption{\small{Overview of our method. We learn occlusion boundaries in a pairwise edgelet MRF framework using unary and continuity occlusion boundary probabilities using edgelet, flow, and geometric features. Then we temporally aggregate the frame by frame predictions to remove inconsistent boundaries.}}
\vspace{-0.5cm}
\label{fig:flowchart}
\end{figure}

\vspace{-0.2cm}
\noindent We provide an overview of our proposed method as shown in \Cref{fig:flowchart}. We begin by over-segmenting the video into spatio-temporal super-voxels. Then, for each neighboring region pair, we extract features to characterize the edgelet between those regions. In particular, we leverage geometric context features to consider the semantic layout in occlusion boundary detection. First, we train a binary classifier to estimate the probability for an edgelet to lie on an occlusion boundary. Next, we enforce occlusion boundary continuity in MRF model by using pairwise edgelet occlusion boundary probability learned by a separate classifier. Finally, we perform temporal smoothing of these estimated occlusion probabilities by aggregating them across successive frames. We perform detailed experiments to show the importance of geometric context features and temporal smoothing for predicting occlusion boundaries in videos. In the following, we describe each step of our algorithm in detail.


\vspace{-0.2cm}
\subsection{Video Segmentation}
\label{sec:videosegmentation}
\vspace{-0.2cm}
We build our algorithm on the initial boundaries provided by video over-segmentation. The purpose of using video segmentation is to find spatio-temporal regions which are coherent in appearance and motion. We use the video segmentation algorithm (and the related online system) proposed by Grundmann~\etal\cite{MatthiasSegmentation} and its extensions~\cite{Corso2012Evaluation}. There method's over-segmentation gives a large number of spatio-temporal super-voxels, which we use as initial candidates for occlusion boundaries. Classifying occlusions on over-segmentation boundaries has following advantages: (1) provides good candidate locations for occlusion detection; (2) it reduces the complexity of the algorithm by not having to classify individual pixels; (3) by working with super-voxels we can enforces temporal coherence or occlusion boundaries; (4) helps in exploiting temporal redundancy and causality for efficient video processing. Next we develop models for learning occlusion boundaries using these candidate boundaries. 

\vspace{-0.2cm}
\subsection{Features for Occlusion Boundary Prediction}
\label{sec:features}
To train classifiers for occlusion boundary prediction, we compute a variety of features. Features are computed on every frame for each edgelet, \textit{i.e.}, boundary between two regions. An edgelet might span more than one frame, in which case it will contribute to the training data multiple times. For each edgelet, we compute features based on boundary, regions, flow, and geometric context. These features are explained next. 

\vspace{-0.3cm}
\paragraph{Boundary and Region Based Features}
Segmentation boundaries provide good candidate locations for finding occlusion boundaries. Longer boundaries with strong color gradients are more likely to be occlusion boundaries as compared to weak short boundaries, we compute boundary length and smoothness for each edgelet, as suggested by~\cite{hoiem2007occl}. In addition to the boundary features, we also include the color difference of the regions surrounding the edgelet. 
\vspace{-0.3cm}
\paragraph{Optical-flow/Motion Based Features}
Motion estimates may have inconsistencies at the occlusion boundaries due to parallax. To capture this information in our framework we compute optical flow based features at each edgelet. We compute optical flow using the total variation method proposed by Wedel \etal\cite{wedel2009improved}. Flow algorithms have photo-consistency assumption. Therefore, pixels advected from reference frame $I_t$ by estimated flow $F_{t\rightarrow t+1}$ should correspond to the next frame $I_{t+1}$. This assumption breaks down at occlusion boundaries, hence high photo-consistency residual should be indicative of such boundaries~\cite{humayun2011learning,ince2008occlusion}. Residual photo-consistency feature $\feat{PC}$ is computed as
\begin{equation}
\feat{PC}(x) \,=\, \left| I_t(x) - I_{t+1}(x + F_{t\rightarrow t+1}(x)) \right|.
\end{equation}

If the motion of two interacting objects is different, their occlusion boundary will have flow discontinuities. To include flow discontinuities, we compute the flow gradient given by
\begin{equation} 
\feat{TG}{}_{,x} \,=\, \|\bigtriangledown u_x\|, \quad \feat{TG}{}_{,y}  \,=\, \| \bigtriangledown v_y \|.
\end{equation}

Since flow gradient is only computed over two pixels, it is unable to capture statistics over a larger area. To capture these proximal flow discontinuities, we compute the variance of the magnitude of flow $F_{\overrightarrow{\text{mag}}} = \|F_{t\rightarrow t+1}\|$ in a spatial window around a pixel given as
\begin{equation}
\feat{mag}(x) \,=\, \mathrm{\mathbf{E}}\left[\left(F_{\overrightarrow{\text{mag}}}(x_i) - \mathrm{\mathbf{E}}\left[F_{\overrightarrow{\text{mag}}}(x)\right]\right)^2\right], 
\end{equation}
\noindent where $x_i$ are the pixels in the $3 \times 3$ window around pixel $x$ and $\mathbf{E}(\cdot)$ is the expectation function. Another way to check inconsistency in flow is to advect pixels by flow $F_{t\rightarrow t+1}$ and follow them back by flow $F_{t+1\rightarrow t}$, \ie, flow computed from $I_{t+1}$ to $I_t$. If the pixel was not occluded or dis-occluded, \ie, it was far from an occlusion boundary, an accurate flow estimate should bring the pixel back to its starting location in frame $I_t$. We use the $\ell_2$ distance from the starting location as a reverse flow constancy feature,
\begin{equation}
\feat{RC}  \,=\, \| x - \left(x_F' + F_{t+1\rightarrow t}(x_F')\right) \|,
\end{equation}
where $x_F' = \text{round}(x + F_{t+1\rightarrow t}(x))$. We can similarly note the inconsistency in the forward and reverse flow angle. $F_{t\rightarrow t+1}$ and $F_{t+1\rightarrow t}$ are said to be consistent if they are $180^\circ$ apart. Any deviation from this is used as a reverse flow angle consistency feature, which is computed as,
\begin{equation}
\feat{RC}{}_{,\theta} \,=\, \left| \pi - \arccos\left[\frac{F_{t\rightarrow t+1}(x)\cdot F_{t+1\rightarrow t}(x_F')}{F_{\overrightarrow{\text{mag}}}(x)F_{\overleftarrow{\text{mag}}}(x_F')}\right] \right|,
\end{equation}
where $F_{\overleftarrow{\text{mag}}} = \|F_{t+1\rightarrow t}\|$ is the magnitude of the reverse optical flow.

\vspace{-0.3cm}
\paragraph{Geometric Layout Features}
Geometric layout estimate provides strong cues for occlusion boundaries and have been shown to be useful for occlusion reasoning and scene understanding \cite{hoiem2008closing}.  For example, an occlusion boundary should exist between different geometric classes, such as, between sky and vertical class(buildings, trees, etc). To include geometric layout estimate for dynamic video scenes, we use the method proposed by Raza~\etal\cite{Raza2013GCFV}. Their method provides confidence for each pixel belonging to geometric classes, such as sky, ground, static-solid, porous, and movable-objects. We use the most likely geometric label and the difference of the average confidence of each geometric class of neighboring regions as feature for occlusion reasoning.




%% file: mrf_occl.tex
\vspace{-0.2cm}
\subsection{MRF Model}
\label{sec:mrf}

Our goal is to maximize the probability of an edgelet $e$ being an occlusion boundary given the edgelet feature vector, \ie, $P(e=\mbox{Occlusion}|X)$. We can estimate this probability in MRF model as,
\begin{equation}
\label{eq:occl_mrf}
\begin{split}
P(e=\mbox{Occlusion}|X)=\frac{1}{Z}\prod\limits_{n=1}^Ng_n(e_n,X_n)\\
\prod\limits_{m\in Conn.(n)}f_{mn}(e_n,e_m)
\end{split}
\end{equation}
where $g_n(\cdot)$ is the unary probability of an edgelet being an occlusion boundary, and $f_{mn}(\cdot,\cdot)$ is the pairwise term capturing the occlusion probability for an edgelet with relation to its connected edgelets. The unary term $g_n(\cdot)$, the occlusion boundary probability of an edgelet, is computed by training a random forest classifier. Random forest are useful for their performance on learning high dimensional non-linear relationships, while providing feature selection and importance for free~\cite{breiman2001random,saeys2008robust}. We trained random forest with 105 trees, 11 random features per node, and a maximum depth of 35 nodes for a tree. We train the unary classifier with the features from each edgelet of each frame to capture the temporal variations. 

The unary classifier, computes the probability of an individual edgelet to be an occlusion boundary edgelet. To enforce continuity of occlusion boundaries, we train a separate random forest classifier to estimate the pairwise edgelet probability. For continuity classifier, we compute the feature for each edgelet pair by concatenating the unary features of both the individual edgelets. The positive pairwise occlusion boundaries are the examples with both edgelets having ground truth occlusion boundary label "true". 

\begin{figure}[t]
\centering
\vspace{-0.5cm}
\includegraphics[width=\columnwidth]{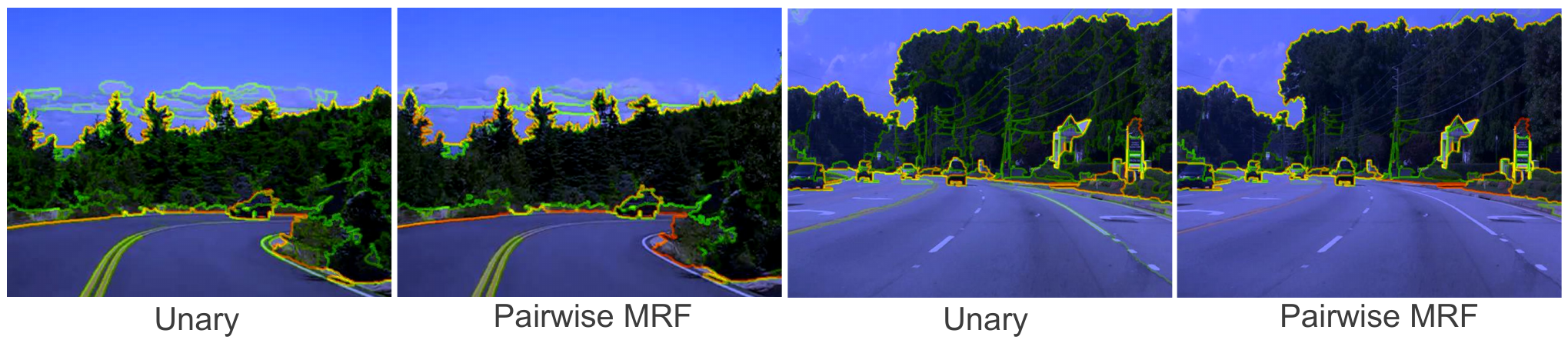}
\caption{\small{Pairwise MRF reduces the false positives in unary prediction as shown above. The yellow, green, and red boundaries show true positives, false positives, and false negatives, respectively.}}
\vspace{-0.4cm}
\label{fig:occl_mrf}
\end{figure}

To predict occlusion boundaries for a test video, we compute occlusion features for each edge of each frame in the over-segmented video. Then we compute the unary and pairwise occlusion boundary probabilities. Final occlusion probability for an edgelet is computed using the MRF model given in \Cref{eq:occl_mrf}. We use loopy-belief propagation algorithm to find the approximate solution for \Cref{eq:occl_mrf}. Pair-wise continuity MRF model reduces false positives over unary occlusion boundary estimate, as shown in \Cref{fig:occl_mrf}. Now, we have assigned each edgelet an occlusion probability and thresholding these probabilities would give occlusion boundaries. However, these estimates may be temporally inconsistent, \textit{i.e.}, occlusion probability of an edgelet may change significantly from one frame to the next. 

To provide temporal consistent occlusion boundaries, we again leverage from video segmentation to temporally smooth the occlusion probability of an edge over a temporal window. The temporal window starts where an edgelet is first formed by two neighboring spatio-temporal regions. Once we have processed the number of instances of a unique spatio-temporal edgelet equal to the length of temporal window, we average the occlusion boundary probabilities in the temporal window for that edgelet, and ignore all future instances of that edgelet. This results in an occlusion boundary algorithm which is linear to the number of unique edgelets in a video than the algorithms which treat video as individual frames and have a complexity of number of edgelets $\times$ number of frames. We experiment with different lengths of temporal windows to filter out temporally inconsistent boundaries (\Cref{sec:results}). 

%% file: results.tex
\vspace{-0.2cm}
\section{Results}
\label{sec:results}
\begin{figure*}
\centering
\vspace{-0.2cm}
\includegraphics[width=0.7\columnwidth]{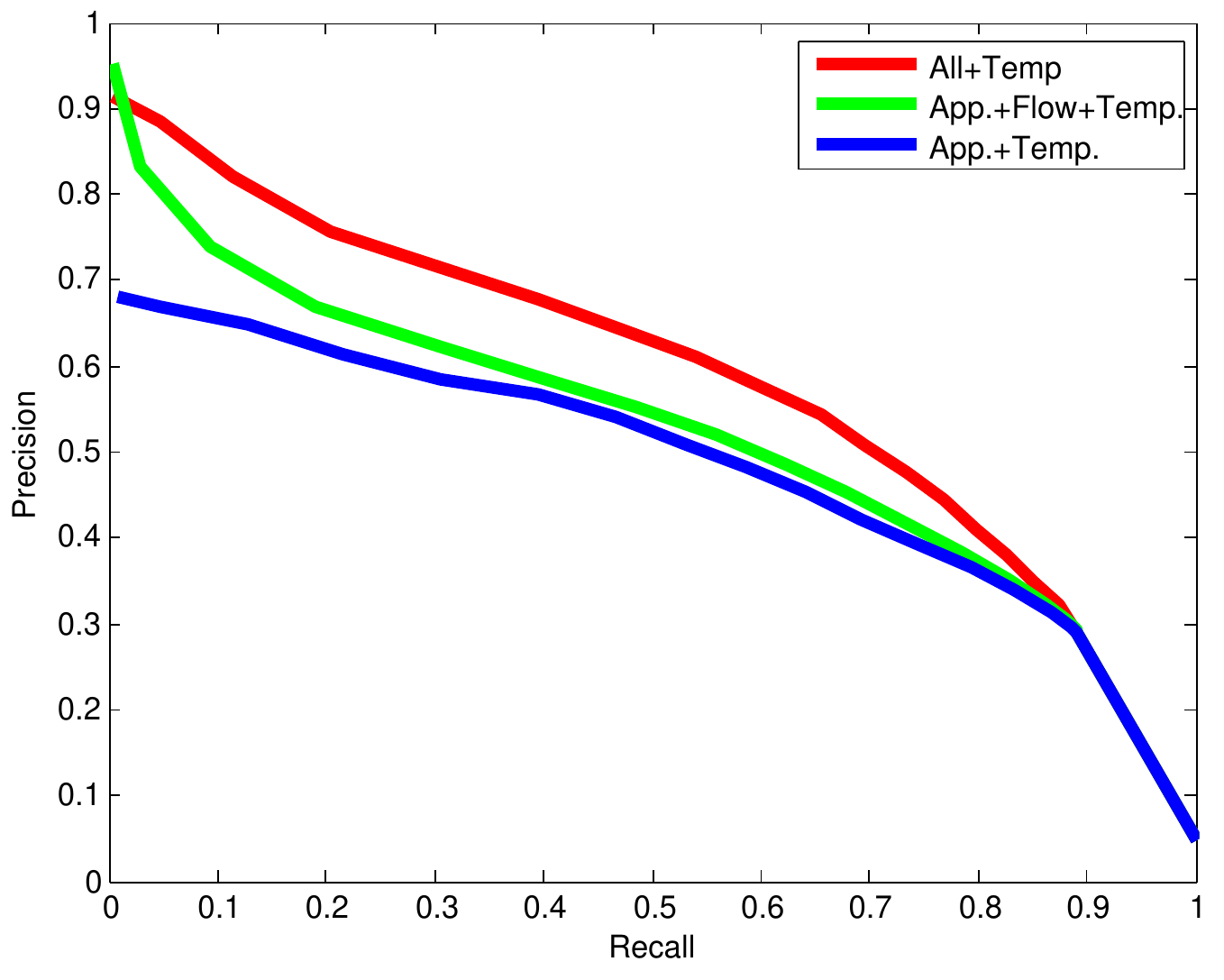}
\hspace{0.1cm}
\includegraphics[width=0.7\columnwidth]{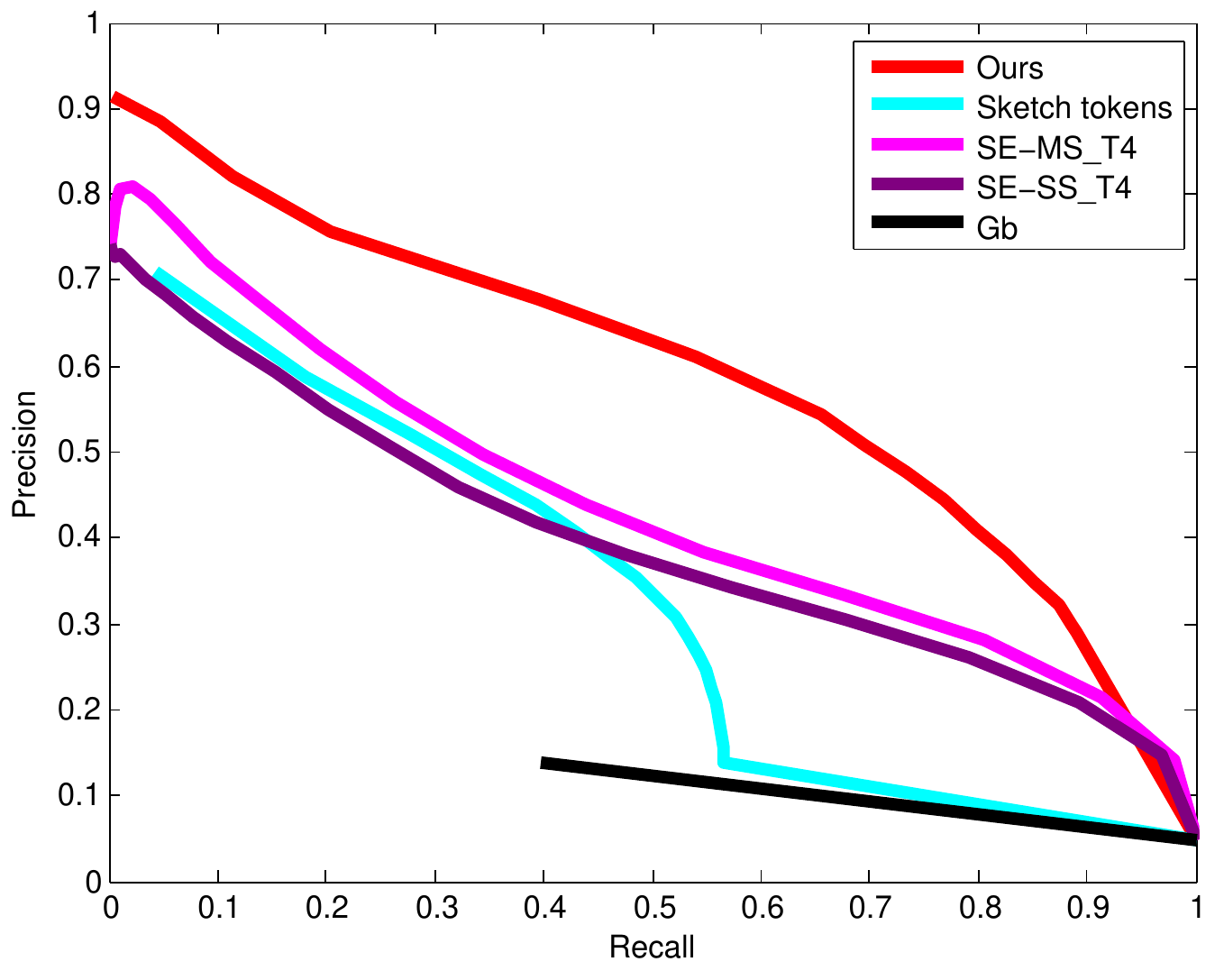}
\caption{\small{Performance evaluation: Precision \emph{vs.} recall (PR) curves for occlusion boundary detection on our dataset. For our algorithm, we used a temporal window of $30$ frames. Legend: ALL (appearance+flow+geometric features), App (appearance features only), and Temp (with temporal smoothing). (Left) Results show that geometric features combined with temporal smoothing yields in the best performance compared to other feature combinations.(Right) Comparison of our method with Sketch Tokens~\cite{lim_CVPR_2013_sketchtokens}, SE-MS\_T4~\cite{dollar_ICCV_2013_edge}, SE-SS\_T4~\cite{dollar_ICCV_2013_edge}, and Gb~\cite{leordeanu2012efficient}.}}
\label{fig:PR_curve}
\vspace{-0.3cm}
\end{figure*}

\vspace{-0.1cm}
In this section, we report the quantitative and qualitative results of our algorithm. Specifically, we measure the performance of our method as precision \emph{vs.} recall  (PR) curves estimated over 5-fold cross-validation by varying the threshold. To compute the precision \emph{vs.} recall curve for our experiments with temporal smoothing, we choose the temporal window with maximum \mbox{F-1} measure. In our experiments, the occlusion boundary prediction performance becomes stable for a temporal window of size greater than $15$ frames. The plot in \Cref{fig:PR_curve} shows that geometric features combined with temporal smoothing results in the best performance. Also, note that temporal smoothing improves performance for each feature set. \Cref{Table:feat_imp} shows \mbox{F-1} measure of each case. 

\begin{figure}
\centering
\includegraphics[width=\columnwidth]{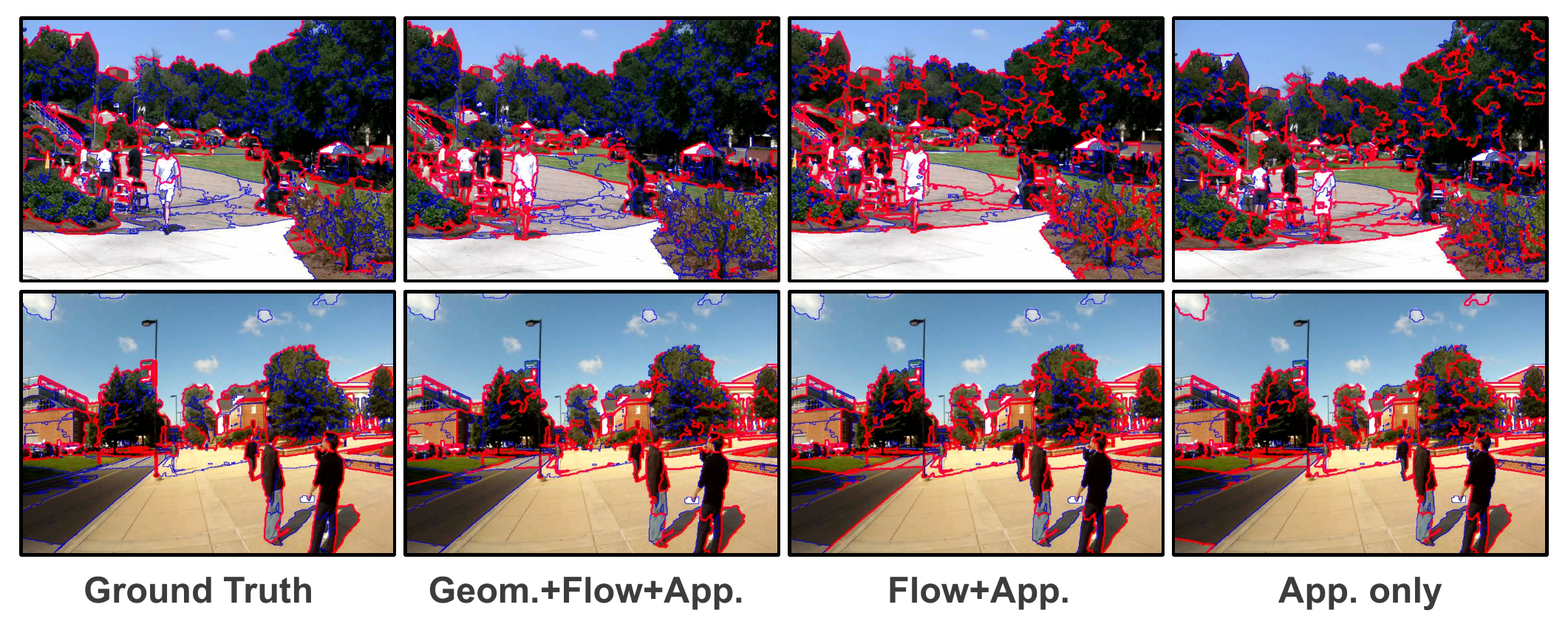}
\caption{\small{Qualitative analysis of feature importance: Figure shows qualitatively that geometric features improve accuracy significantly. Visual comparison is performed with a temporal window of size = 30, and threshold is selected at the peak of \mbox{F-1} measure. Occlusion and non-occlusion boundaries are shown in red and blue, respectively.}}
\vspace{-0.6cm}
\label{fig:feat_imp}
\end{figure}

Our results show qualitative improvement in occlusion boundary detection using geometric context (please see supplementary video). In \Cref{bar:feat_bar}, we show the importance of each feature set from the random forest's out-of-bag training estimate. It is evident from the bar-plot, that geometric features provide more discriminative information for occlusion boundary detection. We show examples to verify the importance of these geometric features, in \Cref{fig:feat_imp}. Note, that the inclusion of geometric features improves occlusion boundary detection by removing boundaries within a geometric class, \eg, boundaries appearing on the ground, across sky, or within trees. Further, they provide important cues to enforce a boundary between different geometric classes.

\begin{table}[t]
  \begin{minipage}{\linewidth}
    \centering
    \small
    \setlength{\tabcolsep}{3pt}
    \begin{tabular}{|l|c|c|c|}\hline
      \textbf{Features} & \textbf{Ind. Frames}  & \textbf{Temporal} \\\hline\hline
      ALL & 0.58  & 0.60 \\\hline
      Appearance+Flow & 0.53  & 0.55 \\\hline
      Appearance Only & 0.52  & 0.53 \\\hline
    \end{tabular}
    \vspace{0.2cm}
    \caption{Comparison of feature sets by \mbox{F-1} measure. Appearance only uses ``Boundary and Region Based Cues''. Appearance+Flow adds to it ``Optical-Flow Based Cues.'' Individual frame-based processing considers each frame individually in the video, whereas the temporal approach takes advantage of causality in videos, by processing over a 30 frame temporal window.} 
    \vspace{0.2cm}
    \label{Table:feat_imp}
  \end{minipage}
  \begin{minipage}{\linewidth}
    \centering
    \small
    \setlength{\tabcolsep}{3pt}
    \begin{tabular}{|l|c|}\hline
      \textbf{Algorithm} & \mbox{\textbf{F-1}} \\\hline\hline
      \textbf{Ours} & \textbf{0.60} \\\hline
      Sketch Tokens~\cite{lim_CVPR_2013_sketchtokens} & 0.42 \\\hline
      SE-MS\_T4~\cite{dollar_ICCV_2013_edge} & 0.46 \\\hline
      SE-SS\_T4~\cite{dollar_ICCV_2013_edge} & 0.43 \\\hline
      Gb~\cite{leordeanu2012efficient} & 0.21 \\\hline
    \end{tabular}
    \vspace{0.2cm}
    \caption{Performance comparison of our method with existing algorithms. Our method exploits causality in videos for temporal occlusion boundary detection.}
    \label{table:algo_comp}
  \end{minipage}
\end{table}

\begin{figure}[t]
\centering
\vspace{-0.2cm}
\includegraphics[width=0.7\columnwidth]{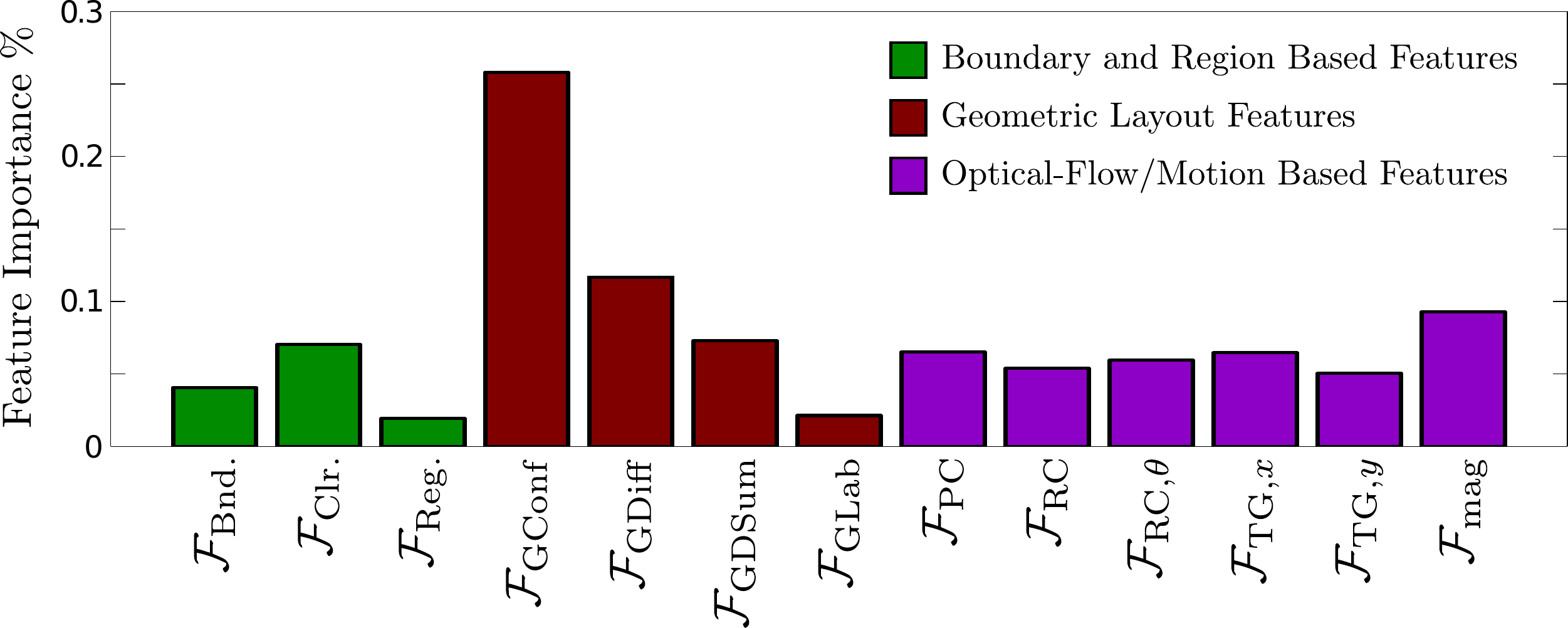}
\caption{\small{Feature importance estimate from random forest over $5$-fold cross-validation. The bar-plot shows the number of votes casted by each feature for the correct class in out-of-bag estimate~\cite{breiman2001random}. Geometric confidence $\feat{GConf}$ estimate of the neighboring regions stands out as most useful along with their difference $\feat{GDiff}$, and the absolute sum $\feat{GDSum}$. Other useful features are flow magnitude variance $\feat{mag}$, photo-consistency $\feat{PC}$, and color feature $\feat{Clr}$. }}
\label{bar:feat_bar}
\vspace{-0.4cm}
\end{figure}

\begin{figure}[htb]
\centering
\vspace{-0.2cm}
\includegraphics[width=\columnwidth]{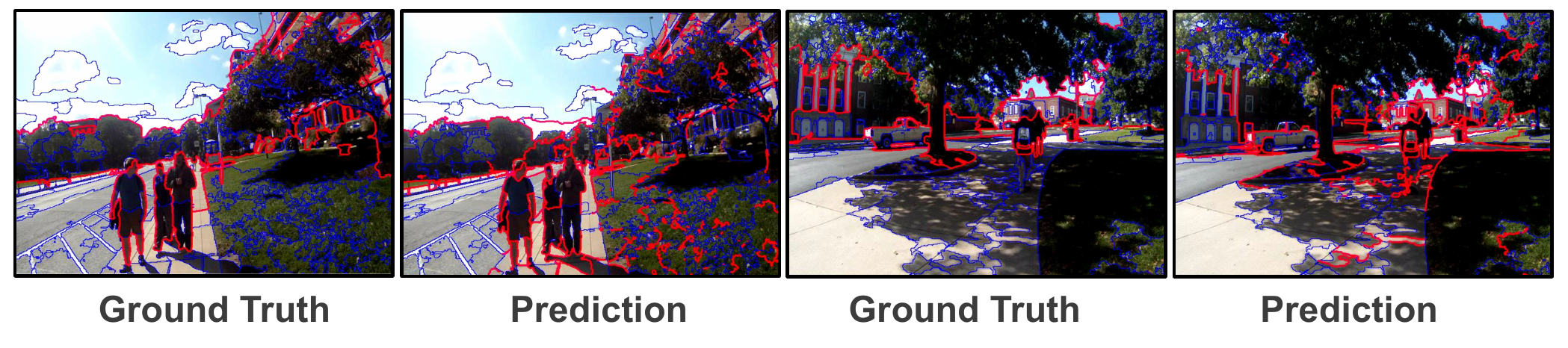}
\caption{\small{Failure cases. Occlusion boundaries are mis-predicted due to shades and fast jittering movement. Temporal smoothing is not useful in fast jittery motion sequences due to short temporal life of segments.}}
\label{fig:misclass_occl}
\vspace{-0.2cm}
\end{figure}

Some misclassification results are shown in \Cref{fig:misclass_occl}. A reason for occlusion boundary misclassification is that we have a very challenging dataset with fast jittery motion. Spatio-temporal segments tend to break quickly in such videos, resulting in very short lived temporal boundaries. For these boundaries temporal smoothing is not effective. Some mis-classifications also occur in shadows due to bad lighting conditions.

\begin{figure}
\centering
\vspace{-0.2cm}
\includegraphics[width=\columnwidth]{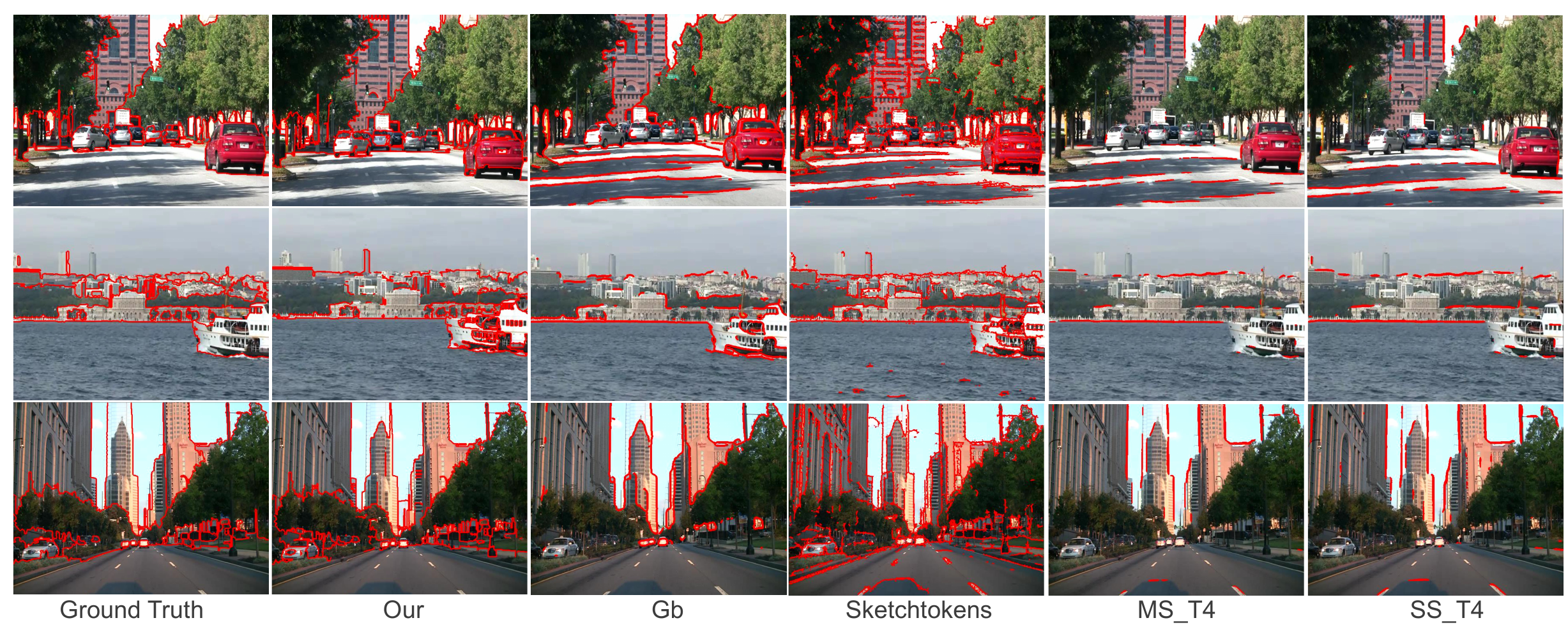}
\caption{\small{Qualitative comparison of occlusion boundaries predicted by \emph{Gb}~\cite{leordeanu2012efficient}, Sketch Tokens~\cite{lim_CVPR_2013_sketchtokens}, and multi-scale (SE-MS\_T4) and single scale (SE-SS\_T4) Structured Edges~\cite{dollar_ICCV_2013_edge}. The probabilistic boundaries are thresholded using the best \mbox{F-1} score over all sequences.}}
\vspace{-0.2cm}
\label{fig:gb}
\end{figure}

\begin{figure*}[t]
\centering
\vspace{-0.2cm}
\includegraphics[width=\textwidth]{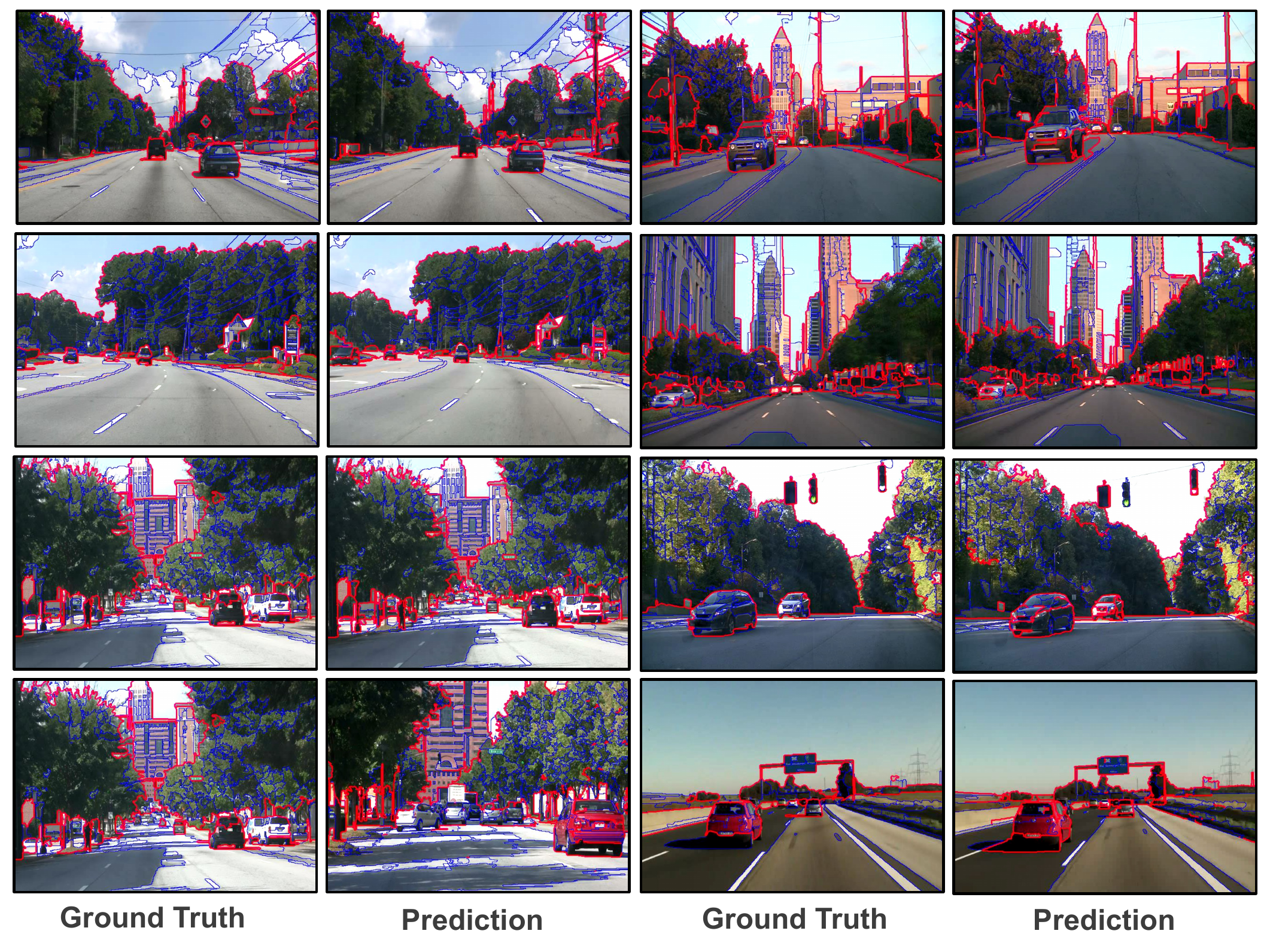}
\caption{\small{Qualitative results for occlusion boundary prediction: (left) Ground truth, (right) Predicted occlusion boundaries using geometric, flow, and appearance features with temporal smoothing (temporal window size=$30$). Occlusion and non-occlusion boundaries are shown in red and blue, respectively.}}
\vspace{-0.2cm}
\label{fig:examples}
\end{figure*}

Direct comparisons and evaluations to other efforts and datasets, with quantitative measures, is hard for our work as there is no such dataset with temporal occlusion boundary, and semantic label annotations (see \Cref{{sec:data}}). In any case, we do undertake and provide a comparison with the occlusion boundaries detected with other occlusion boundary detection algorithms\cite{leordeanu2012efficient,lim_CVPR_2013_sketchtokens,dollar_ICCV_2013_edge}. We applied their publicly available code on our dataset. 
\Cref{table:algo_comp} and \Cref{fig:PR_curve} (Right) shows the comparison of our method with the existing algorithms. To compensate for occlusion boundaries detected by different algorithms in proximity of our ground-truth, we dilate our boundary labelling by a pixel (\ie, an error margin of $3$ pixels). We achieve better performance as compared to other methods. Our algorithm can avoid making false-positive detection within a geometric class, \eg, within tree regions, or boundaries on the ground but other algorithms lack this ability. In addition, by leveraging spatio-temporal occlusion boundaries, we can learn features from all the temporal samples of occlusion boundaries. \Cref{fig:gb} shows qualitative comparison of the above comparison by overlaying occlusion boundaries threholded at best \mbox{F-1} score. It shows that our algorithm can detect occlusion boundaries between different geometric classes, and avoid false positives within a geometric class. \emph{Sketch tokens} algorithm detects most of the boundaries as occlusion boundaries, while \emph{Gb}, \emph{SE-MS\_T4}, and \emph{SE-SS\_T4} detect less boundaries with very few false positives. It should be noted that in our temporal occlusion boundary detection approach, we exploit causality to process the videos efficiently. Temporal occlusion boundary detection only requires $T$ (\ie, length of temporal window) samples of each unique boundary but other approaches require processing the whole video sequence. Operating on a temporal window makes it possible for our algorithm to be be applied to streaming video approaches. \Cref{fig:examples} shows more qualitative results of our approach.




%% file: conclusion.tex
\vspace{-0.2cm}
\section{Conclusion}
\label{sec:conc}
\vspace{-0.1cm}
We have presented an approach for finding temporally consistent occlusion boundaries in dynamic outdoor scenes. We learn occlusion boundaries using edge, flow, and geometric context based features in a pairwise edgelet continuity MRF model. The results are computed on the spatio-temporal boundaries provided by over-segmentation \cite{MatthiasSegmentation}. We choose graph-base video segmentation algorithm for its accuracy in preserving occlusion boudaries, temporal coherence, and ability to handle long video sequences efficiently. However, our approach for learning occlusion boundaries is independent of any particular video segmentation algorithm and should perform well using other video over-segmentation algorithms. The results in this study demonstrate the importance and benefit of integrating scene layout for occlusion reasoning. Moreover, we show that temporal smoothing improves accuracy over independent frame-by-frame processing. Our proposed algorithm also processes videos efficiently by exploiting causality and temporal redundancy using spatio-temporal video segmentation. We have also developed a comprehensive dataset with ground truth temporal occlusion boundary annotations and a broad set of examples containing dynamic scenes. In the future, we plan to integrate more semantic classes and depth information in our method.

\vspace{-0.3cm}
\paragraph{Acknowledgement}
\small
This material is based in part on research by the Defense Advanced Research Projects Agency (DARPA) under Contract No. W31P4Q-10-C-0214, and by a Google Grant and and Google PhD Fellowship for Matthias Grundman, who participated in this research as a Graduate Student at Georgia Tech. Any opinions, findings and conclusions or recommendations expressed in this material are those of the authors and do not necessarily reflect the views of any of the sponsors funding this research.
\vspace{-0.1cm}